\documentclass[sigconf]{acmart}

\acmSubmissionID{139}

\AtBeginDocument{%
  \providecommand\BibTeX{{%
    \normalfont B\kern-0.5em{\scshape i\kern-0.25em b}\kern-0.8em\TeX}}}

\usepackage{graphicx}
\graphicspath{{figure}, {images}, {example}}
\DeclareGraphicsExtensions{.png,.PNG,.jpeg,.JPEG,.jpg,.JPG,.bmp,.BMP}
\usepackage{subcaption}
\usepackage{float}
\usepackage[justification=raggedright]{caption}	
\usepackage{lscape} 
\usepackage{overpic}

\usepackage[lined,ruled,linesnumbered]{algorithm2e}

\SetAlFnt{\small}
\SetAlCapFnt{\small}
\SetAlCapNameFnt{\small}
\SetAlCapHSkip{0pt}
\IncMargin{-\parindent}

\usepackage{booktabs}  
\usepackage{multirow}
\usepackage{enumitem}

\usepackage{mathtools}
\usepackage{amsmath}
\usepackage{bbm}

\usepackage{units}
\usepackage{color}
\usepackage{colortbl}
\definecolor{pltOrange}{HTML}{ff7f0e}
\definecolor{pltBlue}{HTML}{1f77b4}
\definecolor{evalBronze}{HTML}{a77144}
\definecolor{evalSilver}{HTML}{a7a7ad}
\definecolor{evalGolden}{HTML}{d6af36}
\usepackage{pifont}
\def\etal{et al.~}			  

\newcommand{\figref}[1]{Fig.~\ref{fig:#1}} 
\newcommand{\tabref}[1]{Table~\ref{tab:#1}}
\newcommand{\eqnref}[1]{Eqn.~\ref{eq:#1}}

\def\samplePos{\textbf{x}}
\def\refractiveIdx{n}
\def\gradientIdx{\nabla \textbf{n}}
\def\arcLength{s}
\def\stepSize{\Delta s}

\def\near{N}
\def\far{F}
\def\coarse{c}
\def\fine{f}

\def\rayOrigin{\textbf{o}}
\def\sampleRadiance{\textbf{c}}
\def\sampleDensity{\sigma}
\def\accTrans{T}  
\def\adjDist{\delta}
\def\sampleDir{\textbf{d}}  
\def\sampleDist{t}
\def\pixelColor{C}
\def\ray{\mathbbm{r}}
\def\coarseNetwork{F_\theta}
\def\fineNetwork{F_\phi}
\def\boundaryNetwork{F_\psi}

\def\nearPlane{\sampleDist_{\near}}
\def\farPlane{\sampleDist_{\far}}
\def\coarseCount{N_{\coarse}}
\def\fineCount{N_{\fine}}
\def\eikonalFactor{N_{e}}

\def\coarseSampleSet{\ray_\coarse = { \{(\samplePos_i, \sampleDir_i)\}}_{i=0}^{N_\coarse}}
\def\coarseSample{\ray_\coarse}
\def\fineSampleSet{\ray_\fine = {\{(\samplePos_i, \sampleDir_i)\}}_{i=0}^{N_\fine}}
\def\fineSample{\ray_\fine}

\def\coarseDist{\mathbbm{t}_{\coarse}}
\def\fineDist{\mathbbm{t}_{\fine}}

\copyrightyear{2022} 
\acmYear{2022} 
\setcopyright{acmcopyright}
\acmDOI{10.1145/3550340.3564234}
\acmConference[SA '22 Technical
Communications]{SIGGRAPH Asia 2022 Technical Communications}{December
6--9, 2022}{Daegu, Republic of Korea}
\acmBooktitle{}
\acmPrice{}
\acmISBN{}

\begin{document}

\title{Sampling Neural Radiance Fields for Refractive Objects}

\author{Jen-I Pan}
\email{alexkeroro86@gmail.com}
\affiliation{%
  \institution{National Tsing Hua University}
  \city{Hsinchu}
  \country{Taiwan}}

\author{Jheng-Wei Su}
\email{jhengweisu@gapp.nthu.edu.tw}
\affiliation{%
  \institution{National Tsing Hua University}
  \city{Hsinchu}
  \country{Taiwan}}

\author{Kai-Wen Hsiao}
\email{kevin30112@gmail.com}
\affiliation{%
  \institution{National Tsing Hua University}
  \city{Hsinchu}
  \country{Taiwan}}


\author{Ting-Yu Yen}
\email{tingyus995@gmail.com}
\affiliation{%
  \institution{National Tsing Hua University}
  \city{Hsinchu}
  \country{Taiwan}}

\author{Hung-Kuo Chu}
\email{pigjohn@gmail.com}
\affiliation{%
  \institution{National Tsing Hua University}
  \city{Hsinchu}
  \country{Taiwan}}



\begin{abstract}
Recently, differentiable volume rendering in neural radiance fields (NeRF) has gained a lot of popularity, and its variants have attained many impressive results.
However, existing methods usually assume the scene is a homogeneous volume so that a ray is cast along the straight path.
In this work, the scene is instead a heterogeneous volume with a piecewise-constant refractive index, where the path will be curved if it intersects the different refractive indices.
For novel view synthesis of refractive objects, our NeRF-based framework aims to optimize the radiance fields of bounded volume and boundary from multi-view posed images with refractive object silhouettes.
To tackle this challenging problem, the refractive index of a scene is reconstructed from silhouettes.
Given the refractive index, we extend the stratified and hierarchical sampling techniques in NeRF to allow drawing samples along a curved path tracked by the Eikonal equation.
The results indicate that our framework outperforms the state-of-the-art method both quantitatively and qualitatively, demonstrating better performance on the perceptual similarity metric and an apparent improvement in the rendering quality on several synthetic and real scenes.
%
\end{abstract}



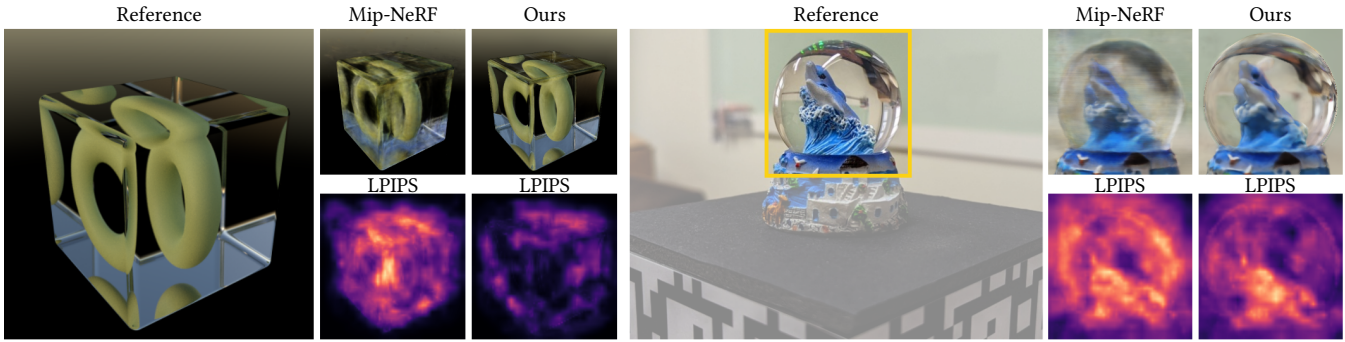
\begin{teaserfigure}
  \begin{overpic}[width=\textwidth,percent]{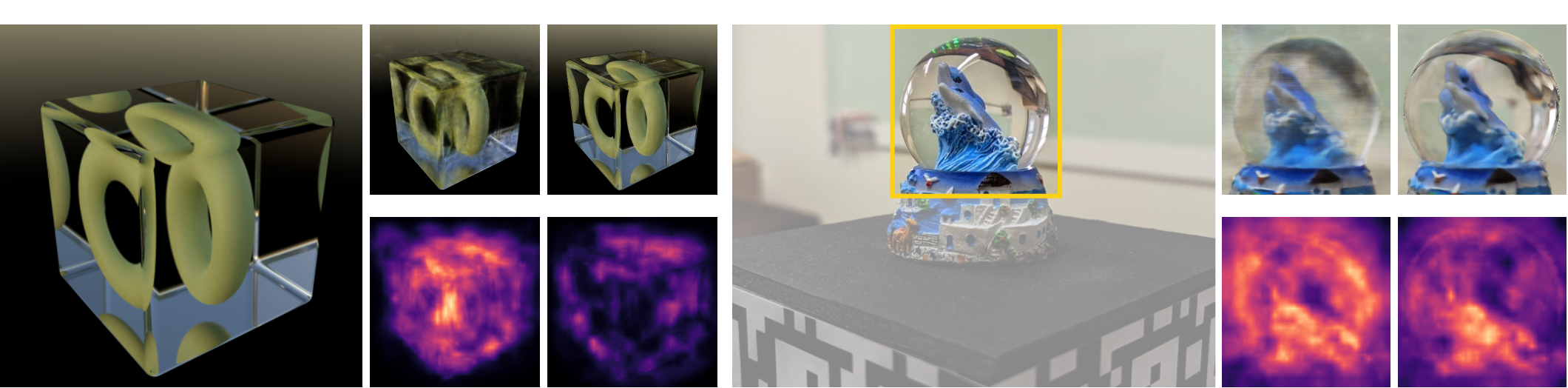}
  \put(11.56, 23.77){\makebox[0pt]{\small Reference}}
  \put(29.02, 23.77){\makebox[0pt]{\small Mip-NeRF}}
  \put(40.35, 23.77){\makebox[0pt]{\small Ours}}
  \put(29.02, 11.00){\makebox[0pt]{\small LPIPS}}
  \put(40.35, 11.00){\makebox[0pt]{\small LPIPS}}
  \put(62.14, 23.77){\makebox[0pt]{\small Reference}}
  \put(83.34, 23.77){\makebox[0pt]{\small Mip-NeRF}}
  \put(94.57, 23.77){\makebox[0pt]{\small Ours}}
  \put(83.34, 11.00){\makebox[0pt]{\small LPIPS}}
  \put(94.57, 11.00){\makebox[0pt]{\small LPIPS}}
  \end{overpic}
  \caption{Our framework takes multi-view images as inputs and renders novel views of both synthetic (left) and real (right) scenes containing refractive objects. With the benefit of considering refraction paths, our results on the surfaces (cube and sphere) and the interior objects (torus and dolphin) are more accurately rendered, as shown in the error maps using LPIPS index (brighter regions indicate higher errors).}
  \label{fig:teaser}
\end{teaserfigure}

\maketitle

\section{Introduction and Related Work}
\label{sec:intro_related}
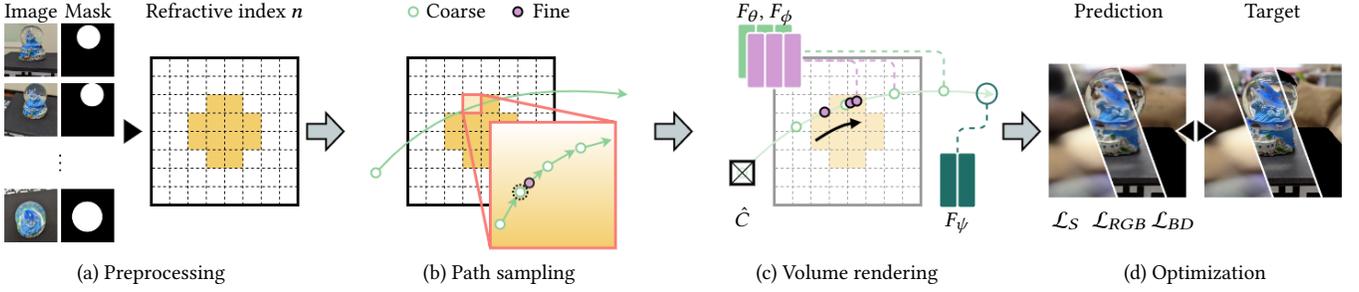
\begin{figure*}[t]
\centering
\begin{overpic}[width=\textwidth]{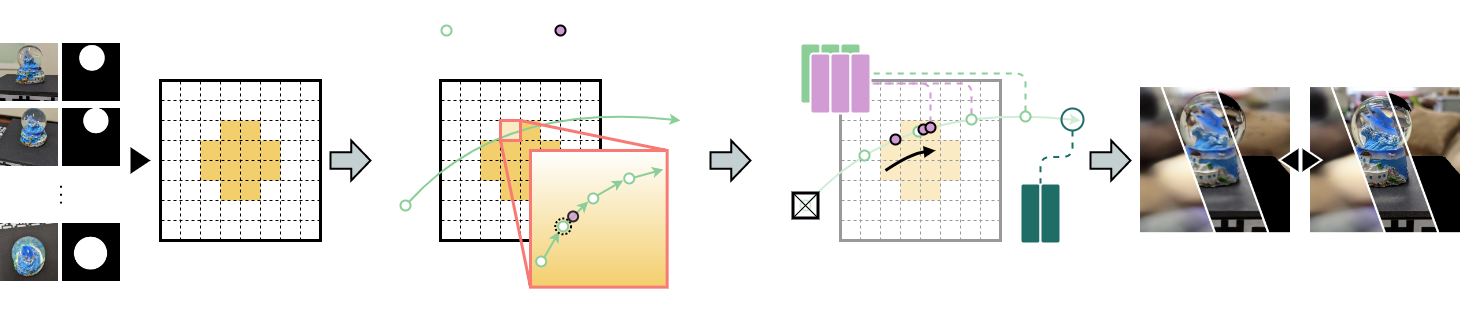}
\put(10.94,0.0){\makebox[0pt]{\small (a) Preprocessing}}
\put(36.94,0.0){\makebox[0pt]{\small (b) Path sampling}}
\put(62.93,0.0){\makebox[0pt]{\small (c) Volume rendering}}
\put(88.92,0.0){\makebox[0pt]{\small (d) Optimization}}
\put(1.98, 19.53){\makebox[0pt]{\small Image}}
\put(6.23, 19.53){\makebox[0pt]{\small Mask}}
\put(16.43, 19.53){\makebox[0pt]{\small Refractive index $\refractiveIdx$}}
\put(56.81, 19.53){\makebox[0pt]{\small $\coarseNetwork$, $\fineNetwork$}}
\put(71.18, 3.97){\makebox[0pt]{\small $\boundaryNetwork$}}
\put(55.1, 3.97){\makebox[0pt]{\small $\hat{\pixelColor}$}}
\put(83.24, 19.53){\makebox[0pt]{\small Prediction}}
\put(94.72, 19.53){\makebox[0pt]{\small Target}}
\put(31.49, 19.53){\small Coarse}
\put(39.49, 19.53){\small Fine}
\put(79.24, 3.97){\makebox[0pt]{\small $\mathcal{L}_S$}}
\put(83.24, 3.97){\makebox[0pt]{\small $\mathcal{L}_{RGB}$}}
\put(87.24, 3.97){\makebox[0pt]{\small $\mathcal{L}_{BD}$}}
\end{overpic}
\caption{
Framework overview. Given the multi-view posed images and refractive object silhouettes, we first reconstruct the refractive index $\refractiveIdx$ of a scene from silhouettes and store it in a voxel grid (a). Next, we track the ray of a pixel $\pixelColor$ and draw the samples along the traversed path (b). Then, we query the density and radiance of each sample from the networks, namely $\coarseNetwork$ and $\fineNetwork$, and combine them with the boundary radiance evaluated by the boundary network $\boundaryNetwork$ to estimate the resulting color (c). Finally, we optimize the three networks with respect to the re-rendering error ($\mathcal{L}_{RGB}$) and regularizers ($\mathcal{L}_{S}$ and $\mathcal{L}_{BD}$) (d).  
}
\label{fig:framework}
\end{figure*}

Refraction is ubiquitous in everyday life. For example, distorted objects seen through the water, and a magnifying glass decreasing the field of view.
%
Thus, accurate rendering of refraction is crucial to improve realism.
Nevertheless, reconstructing the scene with a refractive object from multi-view images is an ill-posed problem due to the ambiguity among geometry, material and refractive index.

For the past two years, neural radiance field, or NeRF~\cite{mildenhall2020nerf}, and its variants that treat a scene as a homogeneous volume have been widely explored.
%
NeRF uses two multi-layer perceptrons (MLPs), one coarse $\coarseNetwork$ and one fine $\fineNetwork$, to represent a volumetric scene. The MLP takes the encoded position $\samplePos$ of a sample and view direction $\sampleDir$ by positional encoding as inputs; it outputs the density $\sampleDensity$ and radiance $\sampleRadiance$.
The pixel value $\hat{\pixelColor}$ is estimated by the differentiable volume rendering equation in \eqnref{volrend} with all the samples $\ray$ along a ray cast from the camera origin $\rayOrigin$ to the pixel.
\begin{align}
\label{eq:volrend}
\begin{split}
& \hat{\pixelColor}(\ray) = \sum_{i=1}^{N} \accTrans_i \left( 1 - \text{exp} \left( -\sigma_i \delta_i \right) \right) \sampleRadiance_i
\end{split} \text{,}
\end{align}
where $\accTrans_i=\text{exp} \left( -\sum_{j=1}^{i-1} \sigma_j \delta_j \right)$ is the accumulated transmittance, and $\adjDist$ is the distance between adjacent samples.
Although NeRF can replace the background with white via the mask of an opaque object, the background seen through a refractive object cannot be easily removed. Thus, the original equation in NeRF (\eqnref{volrend}) is insufficient for our problem due to the lack of a boundary term.
The samples used to estimate the pixel value consist of two sets, one coarse $\coarseSample$ and one fine $\fineSample$, for the coarse and fine networks. Specifically, the fine network takes the union of both sets after sorting. The coarse samples $\ray_\coarse = { \{(\samplePos_i, \sampleDir)\}}_{i=0}^{N_\coarse}$ are drawn uniformly from the evenly-spaced bins between the near $\nearPlane$ and far $\farPlane$ planes with a stratified sampling of $\sampleDist_i$ in \eqnref{stratified} for a continuous representation; the fine samples $\ray_\fine = { \{(\samplePos_i, \sampleDir)\}}_{i=0}^{N_\coarse}$ are then allocated to visible regions that most likely contribute to the pixel value based on the coarse network with a hierarchical sampling of $\sampleDist_i$ in \eqnref{hierarchical}, respectively by the ray equation $\samplePos_i = \rayOrigin + \sampleDist_i \sampleDir$.
%
\begin{align}
\label{eq:stratified}
& \coarseDist = \left\{ \sampleDist_{i} \sim \mathcal{U} \left[ (i-1) / \coarseCount, i / \coarseCount \right] \cdot \left( \farPlane - \nearPlane \right) + \nearPlane \right\}_{i=1}^{\coarseCount}
\end{align}
\begin{align}
\label{eq:hierarchical}
\begin{split}
& \fineDist = \left\{ \sampleDist_{i} \sim \text{InverseTransformSampling} \left( \hat{w}_i \right) \right\}_{i=1}^{\fineCount}
\end{split}
\end{align}
where $\hat{w}_i=w_i/\sum_{j=1}^{\coarseCount} w_j$, and $w_i = \accTrans_i \left( 1 - \text{exp} \left( -\sigma_i \delta_i \right) \right)$.
%
However, these two sampling techniques in NeRF cannot be used with only the distance $\sampleDist$ and view direction $\sampleDir$ if a path is curved due to the refraction.
%
To this end, we combine light transport simulation based on
the \emph{Eikonal equation} with NeRF for the refraction, and we
extend the original sampling techniques in NeRF to curved paths.

In terms of the rendering quality and refraction, mip-NeRF~\cite{barron2021mip} proposes an integrated positional encoding
and achieves the best quality for both single- and multi-scale contents. Ref-NeRF~\cite{verbin2021ref} uses a concept of environment mapping to enable a sharp view-dependent effect. CompLum~\cite{zhu2021neural} brings a surface light field
to avoid the costly evaluation of light paths inside the complex refractive geometry.
Still, none of them takes refraction paths into account.
Furthermore, Matusik~\etal~\shortcite{matusik2002acquisition} introduce an image-based rendering method with the 3D scanner for refractive objects.

Moreover, instead of a bounded scene, a boundary can be treated as different representations~\cite{zhang2020nerf++, hao2021gancraft}.
%
To handle the leftover transmittance or density, GANcraft~\cite{hao2021gancraft} proposes a regularizer.
%
Overall, we represent the boundary as a skybox and bring a proper regularizer to resolve the ambiguity of color blending between refractive object and boundary.
Supplementary materials, codes, and datasets are released for academic usage at \url{https://github.com/alexkeroro86/SampleNeRFRO}.


In summary, we make the following contributions:
\begin{itemize}
\item 
A NeRF-based framework for generating high-quality novel view rendering of refractive objects presenting the refraction and total reflection effects. 

\item A tailor-made hierarchical path sampling technique for 
both straight and curved paths.
\end{itemize}

\paragraph{Concurrent work}
Eikonal Fields~\cite{bemana2022eikonal} aims at the same problem as ours.
Compared to the concurrent work, both methods follow the ray equation of geometric optics derived from the Eikonal equation for a volumetric scene by Eikonal Rendering~\cite{ihrke2007eikonal}. We use the piecewise-linear approximation in~\cite{sun2008interactive} (\eqnref{eikonal}) to construct refraction paths:
%
\begin{align}
\begin{split}
\label{eq:eikonal}
\samplePos_{i+1} = \samplePos_i + \frac{\stepSize}{\refractiveIdx}v_i \text{,}\quad v_{i+1} = v_i + \stepSize \gradientIdx
\end{split} \text{,}
\end{align}
where $\stepSize$ is step size, $v$ is defined by $\refractiveIdx \frac{d\samplePos}{d\arcLength}$, $\refractiveIdx$ is refractive index and $\gradientIdx$ is gradient index.
For the refractive index, Eikonal Fields tackles the challenging task of reconstructing the refractive index of a scene by the dedicated multi-stage training strategy.
We assume the object's material is known, hence its corresponding refractive index (e.g., 1.52 for glass and 1.33 for water). In terms of scene complexity, instead of using a bounding box annotation, we provide more complex interior objects inside different refractive objects for both synthetic and real scenes to inspire the follow-up works.
%
%
%
For the sampling, Eikonal Fields only draws samples uniformly between the bounds, but we further propose a hierarchical path sampling technique.
%
%
Besides, NeReF~\cite{wang2022neref} aims to recover the depth and normal of a flat fluid surface for one-time refraction of the last sample using the Snell's law.

\paragraph{Assumption}
Since we consider the refractive index of a refractive object, samples behind the interior object should be occluded eventually. Therefore, a ray does not change its direction when crossing the interior object. In addition, 
we ignore the outer surface of refractive objects, such as the glass in a glass of water.

\section{Method}
\label{sec:method}
\figref{framework} illustrates an overview of our NeRF-based framework.
To be compatible with the Eikonal equation in \eqnref{eikonal}, 
we first reconstruct the proxy geometry of the refractive object by shape from silhouette and remove the noisy components manually. Then, we choose a voxel grid with tri-linear interpolation to represent the refractive index of a scene.
For each vertex, its refractive index is calculated by $A/(A+B) \cdot 1.0+B/(A+B) \cdot \refractiveIdx$, where $A$ and $B$ is the number of
samples within a voxel that are outside and inside the proxy geometry, respectively.
In addition, to eliminate the stair-step artifacts in rendering, we smooth the voxel grid
before compute the gradient index as Sun~\etal~\shortcite{sun2008interactive}.

\begin{figure}[b]
\centering
\begin{overpic}[width=0.75\linewidth,percent]{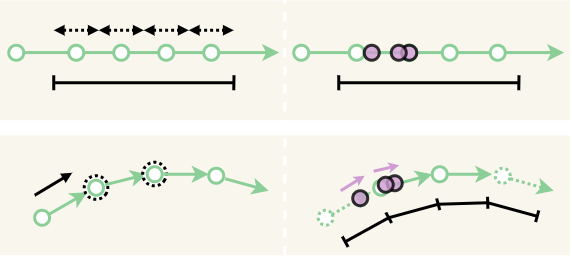}
\put(25, 46.42){\makebox[0pt]{\small Stratified sampling}}
\put(75, 46.42){\makebox[0pt]{\small Hierarchical sampling}}
\put(-3.5, 33){\rotatebox[origin=c]{90}{\small NeRF}}
\put(-3.5, 9.00){\rotatebox[origin=c]{90}{\small Ours}}
\put(9.82, 25.53){\makebox[0pt]{\small $\nearPlane$}}
\put(41.42, 25.53){\makebox[0pt]{\small $\farPlane$}}
\put(13.52, 41.32){\makebox[0pt]{\small $\sampleDist_i$}}
\put(4.88, 41.32){\makebox[0pt]{\small $\rayOrigin$}}
\put(43.86, 41.32){\makebox[0pt]{\small $\sampleDir$}}
\put(59.72, 25.53){\makebox[0pt]{\small $\nearPlane$}}
\put(91.22, 25.53){\makebox[0pt]{\small $\farPlane$}}
\put(65.12, 41.32){\makebox[0pt]{\small $\sampleDist_i$}}
\put(62.86, 17.61){\makebox[0pt]{\small $\sampleDist_i$}}
\end{overpic}
\caption{
Comparison between NeRF's and our sampling techniques. A dotted double arrow is a bin in \eqnref{stratified}.
}
\label{fig:sampling}
\end{figure}
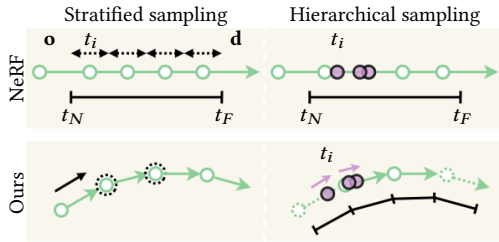
To construct a path according to the refractive index and gradient index, we leverage the Eikonal equation in \eqnref{eikonal} to bend a piecewise-linear path at each step $i$.
We choose the step size $\Delta s=(\farPlane-\nearPlane)/(\coarseCount \times \eikonalFactor)$ to track the path in $\coarseCount * \eikonalFactor$ steps.
Here all the steps are denoted as the \emph{Eikonal samples}, and we also collect the distance of each step.
%
However, if we use all the Eikonal samples to estimate the pixel value, it is costly to evaluate by using the coarse network.
Therefore, we randomly draw one sample in every $\eikonalFactor$ samples to reduce the times of network evaluation, and these drawn samples are called the \emph{coarse samples} $\coarseSampleSet$, where $\sampleDir_i=v_i / || v_i ||$.
For the \emph{fine samples} $\fineSampleSet$, we should not only allocate the fine samples to visible regions as NeRF but also make sure they are still along the piecewise-linear path by extending the hierarchical sampling. With \eqnref{hierarchical}, we transform the set of fine distances $\fineDist$ to the fine samples $\fineSample$ by assigning the direction of the nearest former Eikonal sample $\sampleDir_{\lfloor \sampleDist \rfloor}$ to $\sampleDir_i$ according to the distance $\sampleDist$. Then, we re-calculate the position $\samplePos_i$ based on the position $\samplePos_{\lfloor \sampleDist \rfloor}$ and distance $\lfloor \sampleDist \rfloor$ of the nearest former Eikonal sample along the direction $\sampleDir_i$ by $\samplePos_{\lfloor \sampleDist \rfloor} + \sampleDir_i(\sampleDist - \lfloor \sampleDist \rfloor)$.
We illustrate our sampling techniques and compare to NeRF in \figref{sampling}.

After the network evaluation of samples $\ray$, we gather the corresponding density $\sampleDensity$ and radiance $\sampleRadiance$ by the following volume rendering equation with boundary term:
\begin{align}
\label{eq:volrendBd}
\hat{\pixelColor}(\ray) &= \sum_{i=1}^{N} \accTrans_i (1 - \text{exp}(-\sigma_i \delta_i)) \sampleRadiance_i + \accTrans_{N+1} \pixelColor'(\sampleDir_{N})\text{,}
\end{align}
where $\pixelColor'$ is a skybox represented as a small MLP $\boundaryNetwork: \sampleDir \rightarrow \sampleRadiance$ whose architecture is based on the normal field in NeRFactor~\cite{zhang2021nerfactor}, and $\sampleDir_{N}$ is the leaving direction from the bounded volume.
%

Finally, we optimize the three MLPs, namely $\coarseNetwork$, $\fineNetwork$ and $\boundaryNetwork$, with respect to the following objective function:
\begin{align}
\label{eq:loss}
\begin{split}
\mathcal{L} = \lambda_{RGB} \mathcal{L}_{RGB} + \lambda_{BD} \mathcal{L}_{BD} + \lambda_{S} \mathcal{L}_S
\end{split} \text{,}
\end{align}
where $\lambda_{RGB}$, $\lambda_{BD}$ and $\lambda_{S}$ are the weighted hyper-parameters. We illustrate these terms in \figref{framework}(d).
\paragraph{Re-rendering error}
We use a L2 loss to compare the coarse $\hat{\pixelColor}_{\coarse}$ and fine $\hat{\pixelColor}_{\fine}$ pixel values with the ground truth $C(\textbf{r})$ as NeRF:
\begin{align}
\label{eq:lossRGB}
\begin{split}
\mathcal{L}_{RGB} = \| C(\textbf{r}) - \hat{\pixelColor}_\coarse(\coarseSample) \|_{2}^{2} + \| C(\textbf{r}) - \hat{\pixelColor}_\fine \left( \text{sort}  (\coarseSample \cup \fineSample) \right) \|_{2}^{2}
\end{split} \text{.}
\end{align}

\paragraph{Boundary regularizer}
It is calculated based on the re-rendering error but only updates the density $\sampleDensity$ evaluated by the fine network to preserve the visual quality and eliminate the blurry artifacts on the refractive surface:
\begin{align}
\label{eq:lossBD}
\begin{split}
\mathcal{L}_{BD}=\mathbbm{1}(\accTrans_{\fine, \coarseCount+\fineCount+1}) \cdot \| C(\textbf{r}) - \accTrans_{\fine, \coarseCount+\fineCount+1} \pixelColor'(\sampleDir_{\fine, \coarseCount+\fineCount}) \|_{1}
\end{split} \text{,}
\end{align}
where $\mathbbm{1}(\cdot)$ is an indicator function to ignore the error of occluded region by the threshold $0.5$ on the last accumulated transmittance.

\paragraph{Smoothness regularizer}
We add an L2 gradient penalty to boundary as NeRFactor~\cite{zhang2021nerfactor} on a tile of directions $\sampleDir'$:
\begin{align}
\label{eq:lossS}
\begin{split}
\mathcal{L}_S = \left( 0.5 \cdot \| \begin{bmatrix} -1 & 1 \end{bmatrix} * C'(\sampleDir') \|_2^2 + 0.5 \cdot \| \begin{bmatrix} -1 \\ 1 \end{bmatrix} * C'(\sampleDir') \|_2^2 \right)
\end{split} \text{.}
\end{align}

\section{Result}
\label{sec:result}

\paragraph{Dataset}
We rendered four synthetic scenes, namely \textsc{Ship}, \textsc{Torus}, \textsc{DeerGlobe} and \textsc{StarLamp}, from viewpoints sampled on a full sphere with refraction and total reflection effects. The viewpoints are 100, 100, and 200 views of size $800\times800$ for training, validation, and testing splits, respectively. We resize all the images by half for experiments.
Moreover, we captured one real scene (\textsc{Dolphin}) from viewpoints sampled upon a hemisphere. The viewpoints are 100, 50, and 100 views of size $2560\times1920$ for training, validation, and testing splits, respectively, and the camera poses are calibrated with AprilTag~\cite{krogius2019iros}. We resize all the images by half and crop the center for experiments. 
We also select three real scenes from Eikonal Fields~\cite{bemana2022eikonal}, namely \textsc{Ball}, \textsc{Glass} and \textsc{Pen}, and compare to the provided video sequences.

\begin{figure*}[t]
\centering
\begin{overpic}[width=0.9\textwidth,percent]{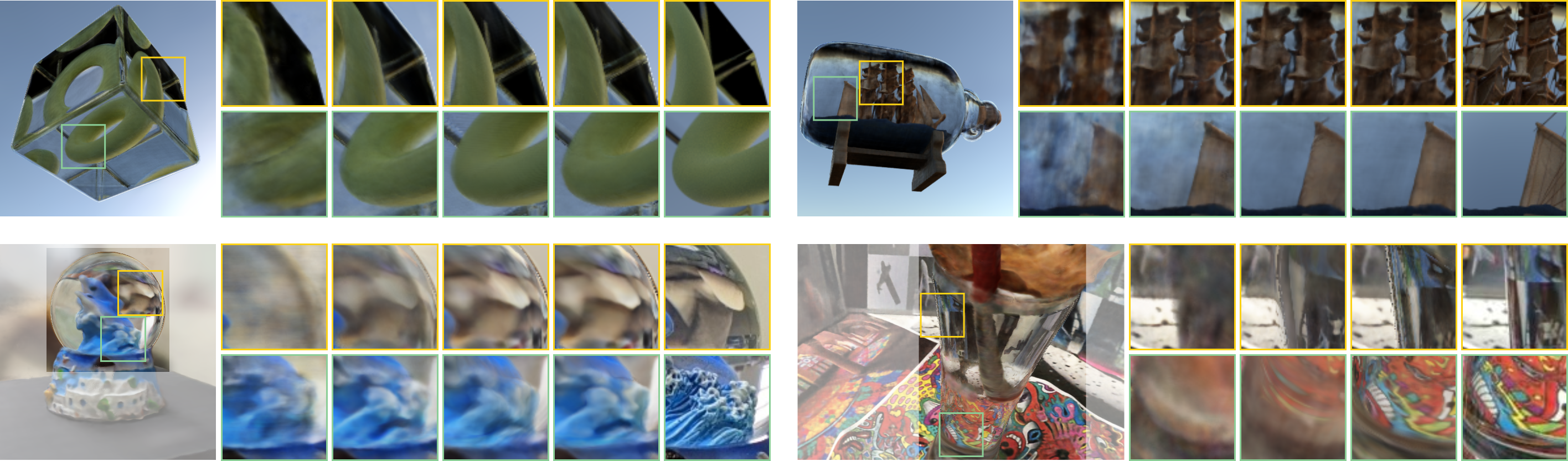}

\put(-1.5, 22.41){\rotatebox[origin=c]{90}{\small \textsc{Torus}}}
\put(-1.5, 6.88){\rotatebox[origin=c]{90}{\small \textsc{Dolphin}}}
\put(49.33, 22.41){\rotatebox[origin=c]{90}{\small \textsc{Ship}}}
\put(49.33, 6.88){\rotatebox[origin=c]{90}{\small \textsc{Pen}}}

\put(6.88, 29.8){\makebox[0pt]{\scriptsize Ours}}
\put(17.47, 29.8){\makebox[0pt]{\scriptsize Mip-NeRF}}
\put(24.53, 29.8){\makebox[0pt]{\scriptsize Ours w/o BD}}
\put(31.59, 29.8){\makebox[0pt]{\scriptsize Ours w/o H}}
\put(38.65, 29.8){\makebox[0pt]{\scriptsize Ours}}
\put(45.71, 29.8){\makebox[0pt]{\scriptsize Reference}}

\put(57.71, 29.8){\makebox[0pt]{\scriptsize Ours}}
\put(68.3, 29.8){\makebox[0pt]{\scriptsize Mip-NeRF}}
\put(75.36, 29.8){\makebox[0pt]{\scriptsize Ours w/o BD}}
\put(82.42, 29.8){\makebox[0pt]{\scriptsize Ours w/o H}}
\put(89.48, 29.8){\makebox[0pt]{\scriptsize Ours}}
\put(96.54, 29.8){\makebox[0pt]{\scriptsize Reference}}

\put(6.88, 14.03){\makebox[0pt]{\scriptsize Ours}}
\put(17.47, 14.03){\makebox[0pt]{\scriptsize Mip-NeRF}}
\put(24.53, 14.03){\makebox[0pt]{\scriptsize Ours w/o BD}}
\put(31.59, 14.03){\makebox[0pt]{\scriptsize Ours w/o H}}
\put(38.65, 14.03){\makebox[0pt]{\scriptsize Ours}}
\put(45.71, 14.03){\makebox[0pt]{\scriptsize Reference}}

\put(61.26, 14.03){\makebox[0pt]{\scriptsize Ours}}
\put(75.36, 14.03){\makebox[0pt]{\scriptsize Mip-NeRF}}
\put(82.42, 14.03){\makebox[0pt]{\scriptsize Eikonal Fields}}
\put(89.48, 14.03){\makebox[0pt]{\scriptsize Ours}}
\put(96.54, 14.03){\makebox[0pt]{\scriptsize Reference}}

\end{overpic}
\caption{
Qualitative comparison of the selected synthetic (top) and real (bottom) scenes. 
}
\label{fig:qualitative}
\end{figure*}
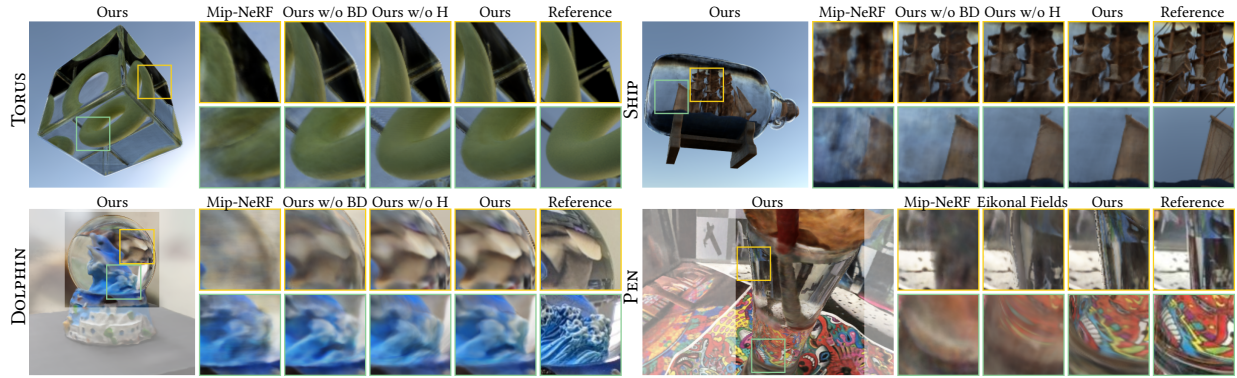

\paragraph{Experimental setting}
%
We choose PSNR and SSIM for low-level image similarity, and LPIPS for better mimicking human preference as our evaluation metrics.
We set $\coarseCount = 64$, $\fineCount = 128$ and 200k training iterations with batch rays 1024 for mip-NeRF~\cite{barron2021mip} and ours. Moreover, during the first 2.5k warm-up iterations, only the re-rendering error in \eqnref{lossRGB} is used.
Note that we crop the object region of an image for evaluating the real scenes.
\paragraph{Ablation study}
We validate our design choices with two experiments on the synthetic scenes. \emph{Ours w/o H} uses no hierarchical sampling but with $\coarseCount=256$. \emph{Ours w/o BD} uses no additional boundary regularizer.
The result in \figref{qualitative} with boundary regularizer shows the better refraction on the surface such as the green box of \textsc{Ship}, and the hierarchical sampling further preserves the details such as the wave in the green box of \textsc{Dolphin}.
%
\paragraph{Competing method}
We compare our method with mip-NeRF~\cite{barron2021mip} and Eikonal Fields~\cite{bemana2022eikonal}.
The results compared with mip-NeRF show that our method achieves a better performance of LPIPS across all the scenes (see~\tabref{quantitative}).
For the comparison against mip-NeRF in \figref{qualitative}, our method preserves much more details (\textsc{Ship}) and generates less blurry results (\textsc{Dolphin}).
%
Then, we compare Eikonal Fields on the selected real scenes.
Our method obtains a comparable LPIPS, and Eikonal Fields cannot reconstruct the refractive index of \textsc{Dolphin} scene (see~\tabref{quantitative}).
%
As shown in \figref{qualitative}, our method could faithfully generate plausible results with better clearness than Eikonal Fields (\textsc{Pen}).
%
%
%

\begin{table}[b]
\caption{
Quantitative comparison of the selected synthetic and real scenes. The top three methods of each metric for a scene are marked by \colorbox{evalGolden!75}{gold}, \colorbox{evalSilver!75}{silver} and \colorbox{evalBronze!75}{bronze}.
}
\label{tab:quantitative}
\addtolength{\tabcolsep}{-3.5pt}

\resizebox{\columnwidth}{!}{
\begin{tabular}{*{13}{lllllllllllll}}
\toprule
& \multicolumn{3}{c}{\textsc{Torus}} & \multicolumn{3}{c}{\textsc{Ship}} & \multicolumn{3}{c}{\textsc{DeerGlobe}} & \multicolumn{3}{c}{\textsc{StarLamp}} \\ \cmidrule(lr){2-4} \cmidrule(lr){5-7} \cmidrule(lr){8-10} \cmidrule(lr){11-13}
& PSNR & SSIM & LPIPS & PSNR & SSIM & LPIPS & PSNR & SSIM & LPIPS & PSNR & SSIM & LPIPS \\ \midrule
Mip-NeRF & 22.32 & 0.759 & 0.268 & 23.93 & \cellcolor{evalBronze!75}{0.828} & \cellcolor{evalBronze!75}{0.151} & 26.79 & 0.881 & 0.134 & \cellcolor{evalSilver!75}{22.04} & \cellcolor{evalGolden!75}{0.885} & \cellcolor{evalBronze!75}{0.092} \\ \midrule
Ours & \cellcolor{evalBronze!75}{25.46} & \cellcolor{evalGolden!75}{0.853} & \cellcolor{evalGolden!75}{0.130} & \cellcolor{evalBronze!75}{24.76} & \cellcolor{evalGolden!75}{0.840} & \cellcolor{evalGolden!75}{0.122} & \cellcolor{evalBronze!75}{27.43} & \cellcolor{evalSilver!75}{0.896} & \cellcolor{evalBronze!75}{0.109} & \cellcolor{evalGolden!75}{22.08} & \cellcolor{evalSilver!75}{0.878} & \cellcolor{evalGolden!75}{0.086} \\
Ours w/o H & \cellcolor{evalSilver!75}{25.57} & \cellcolor{evalSilver!75}{0.852} & \cellcolor{evalBronze!75}{0.136} & \cellcolor{evalSilver!75}{24.77} & \cellcolor{evalSilver!75}{0.838} & \cellcolor{evalSilver!75}{0.127} & \cellcolor{evalSilver!75}{27.58} & \cellcolor{evalBronze!75}{0.894} & \cellcolor{evalSilver!75}{0.108} & \cellcolor{evalGolden!75}{22.08} & \cellcolor{evalBronze!75}{0.876} & \cellcolor{evalSilver!75}{0.091} \\
Ours w/o BD & \cellcolor{evalGolden!75}{25.87} & \cellcolor{evalBronze!75}{0.847} & \cellcolor{evalSilver!75}{0.133} & \cellcolor{evalGolden!75}{25.01} & \cellcolor{evalSilver!75}{0.838} & \cellcolor{evalSilver!75}{0.127} & \cellcolor{evalGolden!75}{30.03} & \cellcolor{evalGolden!75}{0.906} & \cellcolor{evalGolden!75}{0.089} & \cellcolor{evalBronze!75}{21.83} & 0.866 & 0.102 \\
\bottomrule
\end{tabular}
}

\medskip

\resizebox{\columnwidth}{!}{
\begin{tabular}{*{13}{lllllllllllll}}
\toprule
& \multicolumn{3}{c}{\textsc{Dolphin}} & \multicolumn{3}{c}{\textsc{Ball}} & \multicolumn{3}{c}{\textsc{Glass}} & \multicolumn{3}{c}{\textsc{Pen}} \\ \cmidrule(lr){2-4} \cmidrule(lr){5-7} \cmidrule(lr){8-10} \cmidrule(lr){11-13}
& PSNR & SSIM & LPIPS & PSNR & SSIM & LPIPS & PSNR & SSIM & LPIPS & PSNR & SSIM & LPIPS \\ \midrule
Mip-NeRF & \cellcolor{evalGolden!75}{18.48} & \cellcolor{evalGolden!75}{0.459} & 0.479 & \cellcolor{evalBronze!75}{16.29} & \cellcolor{evalSilver!75}{0.523} & \cellcolor{evalBronze!75}{0.418} & \cellcolor{evalGolden!75}{18.44} & \cellcolor{evalGolden!75}{0.475} & \cellcolor{evalBronze!75}{0.414} & \cellcolor{evalGolden!75}{19.13} & \cellcolor{evalSilver!75}{0.487} & \cellcolor{evalBronze!75}{0.428} \\
Eikonal Fields & - & - & - & \cellcolor{evalGolden!75}{18.38} & \cellcolor{evalGolden!75}{0.583} & \cellcolor{evalGolden!75}{0.239} & \cellcolor{evalBronze!75}{17.89} & \cellcolor{evalBronze!75}{0.436} & \cellcolor{evalGolden!75}{0.303} & \cellcolor{evalBronze!75}{18.83} & \cellcolor{evalBronze!75}{0.485} & \cellcolor{evalSilver!75}{0.335} \\ \midrule
Ours & 18.35 & 0.430 & \cellcolor{evalGolden!75}{0.416} & \cellcolor{evalSilver!75}{17.62} & \cellcolor{evalBronze!75}{0.491} & \cellcolor{evalSilver!75}{0.275} & \cellcolor{evalSilver!75}{18.35} & \cellcolor{evalSilver!75}{0.440} & \cellcolor{evalSilver!75}{0.306} & \cellcolor{evalSilver!75}{18.95} & \cellcolor{evalGolden!75}{0.494} & \cellcolor{evalGolden!75}{0.315} \\
\bottomrule
\end{tabular}
}

\end{table}
\section{Discussion and Future Work}
\label{sec:discuss_future}
We present a NeRF-based framework that synthesizes the refraction in novel views and achieves better human perception performance in several scenes. The results show that explicitly tracking curved paths traversing through different refractive indices can produce more visually plausible refraction. Furthermore, with the help of sampling techniques and a boundary regularizer, our framework can further improve surface details and clarity.
%
Our method still has limitations. The blurry geometric details in real scenes result from the imperfect camera poses compared to the synthetic data, and the foggy artifacts appear on refractive surfaces.
In the future, we plan to tackle relighting via environment mapping to enable novel views under a new illumination and optimizing a voxel grid of refractive index to handle more complex refractive objects.




\bibliographystyle{ACM-Reference-Format}
\bibliography{citation}


\begin{thebibliography}{13}


\ifx \showCODEN    \undefined \def \showCODEN     #1{\unskip}     \fi
\ifx \showDOI      \undefined \def \showDOI       #1{#1}\fi
\ifx \showISBNx    \undefined \def \showISBNx     #1{\unskip}     \fi
\ifx \showISBNxiii \undefined \def \showISBNxiii  #1{\unskip}     \fi
\ifx \showISSN     \undefined \def \showISSN      #1{\unskip}     \fi
\ifx \showLCCN     \undefined \def \showLCCN      #1{\unskip}     \fi
\ifx \shownote     \undefined \def \shownote      #1{#1}          \fi
\ifx \showarticletitle \undefined \def \showarticletitle #1{#1}   \fi
\ifx \showURL      \undefined \def \showURL       {\relax}        \fi
\providecommand\bibfield[2]{#2}
\providecommand\bibinfo[2]{#2}
\providecommand\natexlab[1]{#1}
\providecommand\showeprint[2][]{arXiv:#2}

\bibitem[Barron et~al\mbox{.}(2021)]%
        {barron2021mip}
\bibfield{author}{\bibinfo{person}{Jonathan~T Barron}, \bibinfo{person}{Ben
  Mildenhall}, \bibinfo{person}{Matthew Tancik}, \bibinfo{person}{Peter
  Hedman}, \bibinfo{person}{Ricardo Martin-Brualla}, {and}
  \bibinfo{person}{Pratul~P Srinivasan}.} \bibinfo{year}{2021}\natexlab{}.
\newblock \showarticletitle{Mip-nerf: A multiscale representation for
  anti-aliasing neural radiance fields}. In
  \bibinfo{booktitle}{\emph{Proceedings of the IEEE/CVF International
  Conference on Computer Vision}}. \bibinfo{pages}{5855--5864}.
\newblock


\bibitem[Bemana et~al\mbox{.}(2022)]%
        {bemana2022eikonal}
\bibfield{author}{\bibinfo{person}{Mojtaba Bemana}, \bibinfo{person}{Karol
  Myszkowski}, \bibinfo{person}{Jeppe~Revall Frisvad},
  \bibinfo{person}{Hans-Peter Seidel}, {and} \bibinfo{person}{Tobias
  Ritschel}.} \bibinfo{year}{2022}\natexlab{}.
\newblock \showarticletitle{Eikonal Fields for Refractive Novel-View
  Synthesis}.
\newblock \bibinfo{journal}{\emph{arXiv preprint arXiv:2202.00948}}
  (\bibinfo{year}{2022}).
\newblock


\bibitem[Hao et~al\mbox{.}(2021)]%
        {hao2021gancraft}
\bibfield{author}{\bibinfo{person}{Zekun Hao}, \bibinfo{person}{Arun Mallya},
  \bibinfo{person}{Serge Belongie}, {and} \bibinfo{person}{Ming-Yu Liu}.}
  \bibinfo{year}{2021}\natexlab{}.
\newblock \showarticletitle{Gancraft: Unsupervised 3d neural rendering of
  minecraft worlds}. In \bibinfo{booktitle}{\emph{Proceedings of the IEEE/CVF
  International Conference on Computer Vision}}. \bibinfo{pages}{14072--14082}.
\newblock


\bibitem[Ihrke et~al\mbox{.}(2007)]%
        {ihrke2007eikonal}
\bibfield{author}{\bibinfo{person}{Ivo Ihrke}, \bibinfo{person}{Gernot
  Ziegler}, \bibinfo{person}{Art Tevs}, \bibinfo{person}{Christian Theobalt},
  \bibinfo{person}{Marcus Magnor}, {and} \bibinfo{person}{Hans-Peter Seidel}.}
  \bibinfo{year}{2007}\natexlab{}.
\newblock \showarticletitle{Eikonal rendering: Efficient light transport in
  refractive objects}.
\newblock \bibinfo{journal}{\emph{ACM Transactions on Graphics (TOG)}}
  \bibinfo{volume}{26}, \bibinfo{number}{3} (\bibinfo{year}{2007}),
  \bibinfo{pages}{59--es}.
\newblock


\bibitem[Krogius et~al\mbox{.}(2019)]%
        {krogius2019iros}
\bibfield{author}{\bibinfo{person}{Maximilian Krogius}, \bibinfo{person}{Acshi
  Haggenmiller}, {and} \bibinfo{person}{Edwin Olson}.}
  \bibinfo{year}{2019}\natexlab{}.
\newblock \showarticletitle{Flexible Layouts for Fiducial Tags}. In
  \bibinfo{booktitle}{\emph{Proceedings of the {IEEE/RSJ} International
  Conference on Intelligent Robots and Systems {(IROS)}}}.
\newblock


\bibitem[Matusik et~al\mbox{.}(2002)]%
        {matusik2002acquisition}
\bibfield{author}{\bibinfo{person}{Wojciech Matusik},
  \bibinfo{person}{Hanspeter Pfister}, \bibinfo{person}{Remo Ziegler},
  \bibinfo{person}{Addy Ngan}, {and} \bibinfo{person}{Leonard McMillan}.}
  \bibinfo{year}{2002}\natexlab{}.
\newblock \showarticletitle{Acquisition and Rendering of Transparent and
  Refractive Objects}. In \bibinfo{booktitle}{\emph{Proceedings of the 13th
  Eurographics Workshop on Rendering}} (Pisa, Italy)
  \emph{(\bibinfo{series}{EGRW '02})}. \bibinfo{publisher}{Eurographics
  Association}, \bibinfo{address}{Goslar, DEU}, \bibinfo{pages}{267–278}.
\newblock
\showISBNx{1581135343}


\bibitem[Mildenhall et~al\mbox{.}(2020)]%
        {mildenhall2020nerf}
\bibfield{author}{\bibinfo{person}{Ben Mildenhall}, \bibinfo{person}{Pratul~P
  Srinivasan}, \bibinfo{person}{Matthew Tancik}, \bibinfo{person}{Jonathan~T
  Barron}, \bibinfo{person}{Ravi Ramamoorthi}, {and} \bibinfo{person}{Ren Ng}.}
  \bibinfo{year}{2020}\natexlab{}.
\newblock \showarticletitle{Nerf: Representing scenes as neural radiance fields
  for view synthesis}. In \bibinfo{booktitle}{\emph{European conference on
  computer vision}}. Springer, \bibinfo{pages}{405--421}.
\newblock


\bibitem[Sun et~al\mbox{.}(2008)]%
        {sun2008interactive}
\bibfield{author}{\bibinfo{person}{Xin Sun}, \bibinfo{person}{Kun Zhou},
  \bibinfo{person}{Eric Stollnitz}, \bibinfo{person}{Jiaoying Shi}, {and}
  \bibinfo{person}{Baining Guo}.} \bibinfo{year}{2008}\natexlab{}.
\newblock \showarticletitle{Interactive relighting of dynamic refractive
  objects}.
\newblock In \bibinfo{booktitle}{\emph{ACM SIGGRAPH 2008 papers}}.
  \bibinfo{pages}{1--9}.
\newblock


\bibitem[Verbin et~al\mbox{.}(2021)]%
        {verbin2021ref}
\bibfield{author}{\bibinfo{person}{Dor Verbin}, \bibinfo{person}{Peter Hedman},
  \bibinfo{person}{Ben Mildenhall}, \bibinfo{person}{Todd Zickler},
  \bibinfo{person}{Jonathan~T Barron}, {and} \bibinfo{person}{Pratul~P
  Srinivasan}.} \bibinfo{year}{2021}\natexlab{}.
\newblock \showarticletitle{Ref-NeRF: Structured View-Dependent Appearance for
  Neural Radiance Fields}.
\newblock \bibinfo{journal}{\emph{arXiv preprint arXiv:2112.03907}}
  (\bibinfo{year}{2021}).
\newblock


\bibitem[Wang et~al\mbox{.}(2022)]%
        {wang2022neref}
\bibfield{author}{\bibinfo{person}{Ziyu Wang}, \bibinfo{person}{Wei Yang},
  \bibinfo{person}{Junming Cao}, \bibinfo{person}{Lan Xu},
  \bibinfo{person}{Junqing Yu}, {and} \bibinfo{person}{Jingyi Yu}.}
  \bibinfo{year}{2022}\natexlab{}.
\newblock \showarticletitle{NeReF: Neural Refractive Field for Fluid Surface
  Reconstruction and Implicit Representation}.
\newblock \bibinfo{journal}{\emph{arXiv preprint arXiv:2203.04130}}
  (\bibinfo{year}{2022}).
\newblock


\bibitem[Zhang et~al\mbox{.}(2020)]%
        {zhang2020nerf++}
\bibfield{author}{\bibinfo{person}{Kai Zhang}, \bibinfo{person}{Gernot
  Riegler}, \bibinfo{person}{Noah Snavely}, {and} \bibinfo{person}{Vladlen
  Koltun}.} \bibinfo{year}{2020}\natexlab{}.
\newblock \showarticletitle{Nerf++: Analyzing and improving neural radiance
  fields}.
\newblock \bibinfo{journal}{\emph{arXiv preprint arXiv:2010.07492}}
  (\bibinfo{year}{2020}).
\newblock


\bibitem[Zhang et~al\mbox{.}(2021)]%
        {zhang2021nerfactor}
\bibfield{author}{\bibinfo{person}{Xiuming Zhang}, \bibinfo{person}{Pratul~P
  Srinivasan}, \bibinfo{person}{Boyang Deng}, \bibinfo{person}{Paul Debevec},
  \bibinfo{person}{William~T Freeman}, {and} \bibinfo{person}{Jonathan~T
  Barron}.} \bibinfo{year}{2021}\natexlab{}.
\newblock \showarticletitle{Nerfactor: Neural factorization of shape and
  reflectance under an unknown illumination}.
\newblock \bibinfo{journal}{\emph{ACM Transactions on Graphics (TOG)}}
  \bibinfo{volume}{40}, \bibinfo{number}{6} (\bibinfo{year}{2021}),
  \bibinfo{pages}{1--18}.
\newblock


\bibitem[Zhu et~al\mbox{.}(2021)]%
        {zhu2021neural}
\bibfield{author}{\bibinfo{person}{Junqiu Zhu}, \bibinfo{person}{Yaoyi Bai},
  \bibinfo{person}{Zilin Xu}, \bibinfo{person}{Steve Bako},
  \bibinfo{person}{Edgar Vel{\'a}zquez-Armend{\'a}riz}, \bibinfo{person}{Lu
  Wang}, \bibinfo{person}{Pradeep Sen}, \bibinfo{person}{Milo{\v{s}}
  Ha{\v{s}}an}, {and} \bibinfo{person}{Ling-Qi Yan}.}
  \bibinfo{year}{2021}\natexlab{}.
\newblock \showarticletitle{Neural complex luminaires: representation and
  rendering}.
\newblock \bibinfo{journal}{\emph{ACM Transactions on Graphics (TOG)}}
  \bibinfo{volume}{40}, \bibinfo{number}{4} (\bibinfo{year}{2021}),
  \bibinfo{pages}{1--12}.
\newblock


\end{thebibliography}


\end{document}